\documentclass[10pt, a4paper]{article}

\usepackage{lrec-coling2024}

\usepackage{inconsolata}
\usepackage{covington}
\usepackage{xcolor}
\definecolor{myGreen}{RGB}{0,158,115} 
\usepackage{booktabs}
\usepackage{numprint}
\usepackage{multirow}
\usepackage{siunitx}
\usepackage{tipa}
\usepackage{forest}
\usepackage[normalem]{ulem}
\DeclareFontSubstitution{T3}{cmss}{m}{n}
\usepackage{amsmath}
\usepackage[italic]{mathastext}

\title{\textsc{Killkan}: The Automatic Speech Recognition Dataset for Kichwa with Morphosyntactic Information}

\name{Chihiro Taguchi$^1$, Jefferson Saransig$^2$, Dayana Velásquez$^1$, David Chiang$^1$} 

\address{\begin{tabular}{cc}
    $^1$University of Notre Dame & $^2$Pontifical Catholic University of Ecuador \\
    Notre Dame, IN, USA & Quito, Ecuador \\
    \texttt{\{ctaguchi,dvelasqu,dchiang\}@nd.edu} & \texttt{jisaransig@puce.edu.ec}
\end{tabular}
}

\newcommand{\totalnumtokens}{26,544}
\newcommand{\totaltrainsamples}{3,128}
\newcommand{\totalhours}{4}
\newcommand{\totalnumsamples}{3,928}
\newcommand{\avgsamplelen}{3.59}
\newcommand{\avgsentlen}{6.76}
\newcommand{\avgtokenlen}{6.12}

\abstract{
This paper presents \textsc{Killkan}, the first dataset for automatic speech recognition (ASR) in the Kichwa language, an indigenous language of Ecuador.
Kichwa is an extremely low-resource endangered language, and there have been no resources before \textsc{Killkan} for Kichwa to be incorporated in applications of natural language processing.
The dataset contains approximately \totalhours~hours of audio with transcription, translation into Spanish, and morphosyntactic annotation in the format of Universal Dependencies.
The audio data was retrieved from a publicly available radio program in Kichwa.
This paper also provides corpus-linguistic analyses of the dataset with a special focus on the agglutinative morphology of Kichwa and frequent code-switching with Spanish.
The experiments show that the dataset makes it possible to develop the first ASR system for Kichwa with reliable quality despite its small dataset size.
This dataset, the ASR model, and the code used to develop them will be publicly available.
Thus, our study positively showcases resource building and its applications for low-resource languages and their community.
 \\ \newline }

\begin{document}

\maketitleabstract

\section{Introduction}
Language endangerment has been one of the world's cultural crises, by which many of the world's languages are losing their speakers at an unprecedented pace \citep{belew-2018}.
Among efforts to document and revitalize languages, recent years have seen growing attention and work to incorporate digital technologies and media to this end \citep{jimerson-prudhommeaux-2018-asr, michaud-2018, prudhommeaux-2021, shi-etal-2021-leveraging, tsoukala-etal-2023-asr}.
In the same spirit, this study presents the \textsc{Killkan}\footnote{\textsc{Killkan} stands for \textit{\textbf{Ki}chwa uyashkata pay\textbf{ll}atak \textbf{k}illkak \textbf{an}ta} (Kichwa automatic speech recognizer) in Kichwa. The word \textit{killkan} also means ``it writes''.} corpus, the first dataset for automatic speech recognition (ASR) for the Kichwa language.
Though Kichwa is estimated to have a few hundred thousand speakers in Ecuador, it is considered endangered, as the society is undergoing a language shift to Spanish only.
In natural language processing (NLP), Kichwa is an extremely low-resource language, as there have been no datasets available for either building language models or conducting computational linguistic research of Kichwa.

Our dataset consists of \totalhours~hours of audio with its orthographic transcription containing \totalnumtokens~tokens.
Furthermore, each sentence is annotated with its Spanish translation and morphosyntactic information in the CoNLL-U format of Universal Dependencies (UD) \citep{nivre-2020}.
To evaluate the utility of the dataset, we train ASR models on it by fine-tuning the pretrained model \texttt{wav2vec2-xlsr-53}.
The experiment shows that the fine-tuned model's performance was 2.04\%~Character Error Rate (CER), which is comparable to Wav2Vec2 models fine-tuned on high-resource languages.

In the following section, we provide a linguistic overview of the Kichwa language.
Then, Section \ref{sec:related} surveys previous related work done in the field of NLP for Quechuan languages.
Section \ref{sec:dataset} describes the details of our dataset, including the data source, the annotation process, and a brief analysis of the dataset.
Section \ref{sec:experiments} reports the experimental results of training ASR models on our dataset, followed by concluding remarks in Section \ref{sec:conclusion}.

Our contributions in this work are the following:
\begin{itemize}
    \item We publish the first dataset for Kichwa containing manually annotated audio, transcription, its Spanish translation, and morphosyntactic information;
    \item We develop the first ASR models for Kichwa;
    \item We present a new UD Treebank for Kichwa incorporated in ELAN annotation;
    \item Our dataset, the ASR models, and the code used to develop them are publicly available.\footnote{The dataset and the code are available in \url{https://github.com/ctaguchi/killkan}, and the model is available in \url{https://huggingface.co/ctaguchi/killkan_asr}.}
\end{itemize}

\section{Background}\label{sec:background}
\paragraph{The Kichwa language.}
Kichwa is the most widely spoken indigenous language in the Republic of Ecuador, particularly along the Andean mountain range in the middle and the Amazonian region to the east of the country.
Though the number of speakers greatly varies among different statistics, the language is estimated to have at least 300,000 speakers \citep{king-2002}.
Kichwa is classified in the Northern Quechua branch of the Quechua II group in the Quechuan language family.
Though the Quechua II group also includes more widely spoken varieties such as Cuzco Quechua and Ayacucho Quechua of Peru, Kichwa shows a number of differences from them in phonology and morphosyntax.
For example, Kichwa has lost ejective consonants, possessive suffixes, the inclusive/exclusive distinction in the first-person plural pronoun, has a reduced system of evidentiality \citep{adelaar-2021}.

\begin{figure}
    \centering
    \includegraphics[width=\linewidth]{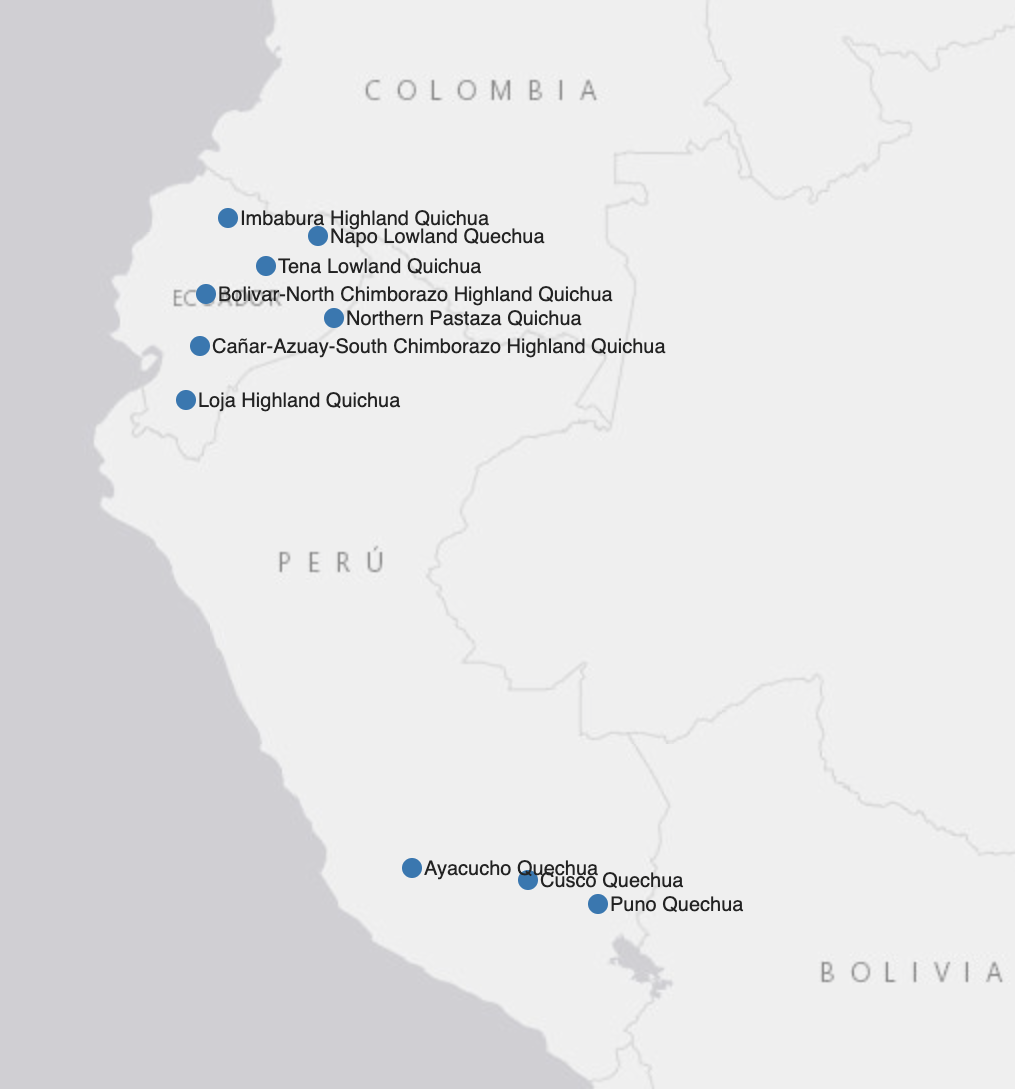}
    \caption{The distribution of Quechua II languages mentioned in this paper.
    This map was created with the \texttt{lingtypology} library \citep{lingtypology}.}
    \label{fig:quechua_map}
\end{figure}

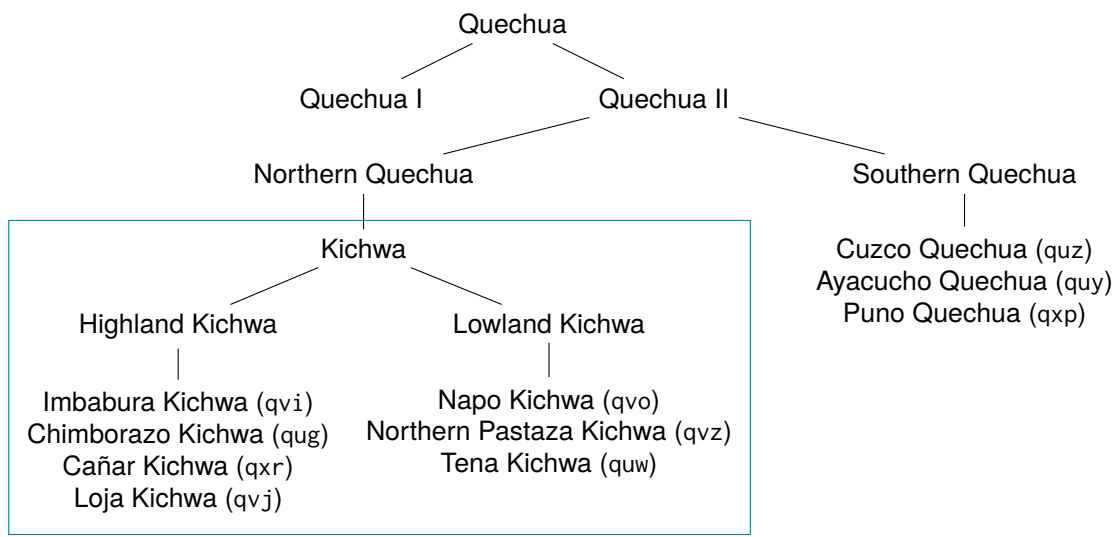
\begin{figure*}
    \centering
    \begin{forest}
        [Quechua, s sep=20mm
            [Quechua I]
            [Quechua II, s sep=20mm
                [Northern Quechua
                    [Kichwa,tikz={\node [draw,teal,fit to=tree] {};}
                        [Highland Kichwa
                            [Imbabura Kichwa (\texttt{qvi}) \\
                            Chimborazo Kichwa (\texttt{qug}) \\
                            Cañar Kichwa (\texttt{qxr}) \\
                            Loja Kichwa (\texttt{qvj}), align=center]]
                        [Lowland Kichwa
                            [Napo Kichwa (\texttt{qvo}) \\
                            Northern Pastaza Kichwa (\texttt{qvz}) \\
                            Tena Kichwa (\texttt{quw}), align=center]]]
                        ]
                [Southern Quechua
                    [Cuzco Quechua (\texttt{quz}) \\
                    Ayacucho Quechua (\texttt{quy}) \\
                    Puno Quechua (\texttt{qxp}), align=center]]]]
    \end{forest}
    \caption{A classification of Quechuan languages with their ISO 639-3 language codes with a focus on the varieties mentioned in this paper.
    The branches in the box are called Ecuadorian Kichwa.}
    \label{fig:classification}
\end{figure*}

Kichwa is in fact an umbrella term that involves several regional varieties of Northern Quechua.
The Endangered Language Project \citeplanguageresource{elp} lists Highland Ecuadorian Kichwa and Lowland Ecuadorian Kichwa, under which several subvarieties are further categorized.
See Figure \ref{fig:classification} for a summary of the classification of Quechuan varieties, and see Figure \ref{fig:quechua_map} for a map of their distribution.

With regard to typological aspects, like other Quechuan varieties, Kichwa is an agglutinative language, where verbal and nominal suffixes and discourse clitics are attached to the root to mark verbal features, cases, and information structure.
The example in (\ref{aggl_ex}) shows the agglutination of voice, tense, case, and topic morphemes on the verb root \textit{llamka} ``to work''.

\begin{covexample}\label{aggl_ex}
    \digloss[]
    {llamka-naku-nka-kaman=ka}
    {work-\textsc{rcp-prosp-ter=top}}
    {Until (someone) works together\footnote{\textsc{rcp}: reciprocal voice, \textsc{prosp}: prospective aspect, \textsc{ter}: terminative case, \textsc{top}: topic.}}
\end{covexample}

\paragraph{Language contact and code-switching.}
Since the arrival of Spanish colonizers in the 16th century, Quechuan languages have had language contact with Spanish \citep{torero-2007}.
The centuries of bilingualism in the Andean region have strongly influenced the lexicon of Quechuan languages, and it is common for Kichwa speakers to code-switch to Spanish in daily speech, which is a language variety sometimes referred to as Media Lengua \citep{deibel-2019}.
The spoken samples in our dataset also contain code-switched speech with Spanish.
The code-switched segment can range from a morpheme (also called intra-word code-switching \citep{nguyen-2016}) to a whole phrase;
examples from the dataset are shown in (\ref{intrawordcs}) and (\ref{phrasalcs}), respectively, where code-switched parts are in green and underlined.

\begin{covexample}\label{intrawordcs}
    \digloss[]
    {Ñuka=rak \uline{\textcolor{myGreen}{vacuna}}-ri-kri-ni.}
    {\textsc{1sg=cont} vaccinate-\textsc{refl-prosp-prs.1sg}}
    {I am going to get vaccinated first.\footnote{\textsc{1sg}: first person singular, \textsc{refl}: reflexive voice, \textsc{prs}: present tense.}}
\end{covexample}

\begin{covexample}\label{phrasalcs}
    \digloss[]
    {\uline{\textcolor{myGreen}{Consulta}} \uline{\textcolor{myGreen}{popular}} alli=mi ri-ku-n.}
    {inquiry popular good=\textsc{foc} go-\textsc{prog-prs.3}}
    {The referendum is going well.\footnote{\textsc{foc}: focus, \textsc{prog}: progressive aspect, \textsc{3}: third person.}}
\end{covexample}

Most Kichwa speakers are bilingual with Spanish and speak Spanish with non-Kichwa speakers.
Though still estimated to have a few hundred thousand speakers, Kichwa is an endangered language that younger generations often do not inherit, speaking only Spanish instead \citep{acosta-2017}.
In addition, Kichwa is both politically and socially marginalized, as suggested by the pejorative term for Kichwa, \textit{yanka shimi} ``useless language'' \citep{maldondo-2007, kowii-2017}.
Unlike Quechua in Peru and Bolivia, Kichwa is merely a recognized language and is not granted an official status in Ecuador.
These factors add to the ongoing endangerment of the language, and resource building for language technologies is indispensable for both documentation and revitalization of the language.
Yet, it is worth mentioning that there are ongoing revitalization activities and Kichwa--Spanish bilingual schooling in Ecuador.

\paragraph{Orthography.}
The orthography of Kichwa is based on the Latin alphabet.
The modern orthographic standardization of Kichwa has undergone two crucial modifications in the late 20th century.
The first attempt to standardize the Kichwa orthography was proposed in 1980.
This orthography exhibits several influences from the Spanish orthography, such as the use of <c> and <q> for the phoneme /k/ (e.g., <quillca> /kilka/ `writing') and the use of <hu> to represent the phoneme /w/ (e.g., <huahua> /wawa/ `child').
In 1998, the orthography was revised again and has been the standard since then \citep{chasiquiza-2019}.
The major modifications are phonology-based simplification, where redundant graphemes such as <qu>/<c> for /k/ and <hu> for /w/ were changed to <k> and <w>, respectively.
Though the old orthography is still sometimes informally used, the transcription in our dataset is in the new orthography since the latter orthography is officially and widely used in today's writing.

The modern Kichwa orthography has 18 letters including three digraphs:
<a>, <ch>, <h>, <i>, <k>, <l>, <ll>, <m>, <n>, <ñ>, <p>, <r>, <s>, <sh>, <t>, <u>, <w>, <y>.
On top of this, two graphemes, <ts> and <z>, may also be used for a small number of words depending on dialects.
For code-switched Spanish words, Spanish orthography is used, though Kichwa orthography may also be used for old loanwords such as <ura> `time' from Spanish \textit{hora}.
Though the correspondence between the orthography and pronunciation is more or less regular, there are slight dialectal differences in the actual phonetic value for each grapheme.
For example, the word \textit{alli} ``good'' is pronounced as /ali/ in Imbabura Kichwa and /a\scalebox{1.2}{\textctz}i/ in Chimborazo Kichwa.

\section{Related Work}\label{sec:related}
Although there is no previous dataset for Ecuadorian Kichwa, there have been several efforts to create datasets and NLP applications for other related Quechuan languages, especially Southern Quechua varieties such as Cuzco Quechua of Peru.
\citet{rios-2014} developed a text normalization pipeline and a morphological analyzer for Cuzco Quechua, to which a machine translation system and a dependency treebank are added in their later work \citep{rios-2016}.
\citet{cardenas-2018} is a speech corpus for Ayacucho (Chanca) Quechua and Puno (Collao) Quechua, which are both Southern Quechuan languages of the Quechua II group spoken in Peru.
\citet{ortega-2020} introduces a new parallel text corpus and trains a neural machine translation system for Quechua from Peru and Bolivia, though it does not mention which specific Quechuan variety the text is written in.
Since Quechuan languages are highly agglutinative, they have been sometimes used in morphology-related tasks in NLP.
For example, \citet{chen-2021} investigates the effect of morphology-aware segmentation instead of Byte-Pair Encoding (BPE) on Quechua.\footnote{
The paper does not mention what variety of Quechua was used in their experiments. Their dataset description implies that some Peruvian varieties were used.
}
More recently, another speech dataset for Peruvian indigenous language was released that includes $\sim$180~hours of Southern Quechua audio \citep{zevallos-etal-2022-huqariq}.
All in all, NLP research and applications for Quechuan languages have centered around the varieties spoken in Peru and Bolivia, and other varieties like Ecuadorian Kichwa have yet to be included in language technologies.

\section{Dataset}\label{sec:dataset}
This section describes the details of our dataset.

\subsection{Source}
The source of the audio in the dataset is a radio program ``Jaboneropak Ayllullaktapi'' (In the neighborhood of Jabonero) provided by Radialistas Apasionadas y Apasionados,\footnote{
\url{https://radialistas.net}.
}
an Ecuador-based non-profit radio station.
It is a compilation of fictional stories related to life during the COVID-19 pandemic.
The program is published under a Creative Commons BY-SA license, permitting re-use and re-distribution of the work.
The acted characters include male and female with various voice qualities and with both adult and child roles.
Though the detailed demographic information of the voice actors is unavailable, it is certain that the speech contains several regional varieties of Highland Kichwa.
The radio program contains 20~episodes in total, each of which has a length of $\sim$12~minutes approximately.
The total audio length of the whole dataset is $\sim$234.86~minutes ($\sim$3.91~hours).
The dataset contains \totalnumsamples~samples, where each audio sample corresponds to a sentence.
The average length of a sample is $\sim$\avgsamplelen~seconds.
The transcription contains \totalnumtokens~tokens, and the average length of a token was $\sim$\avgtokenlen~characters.
The average sentence length was $\sim$\avgsentlen~tokens.

\subsection{Annotation}
\begin{figure*}[t]
    \centering
    \includegraphics[width=\textwidth]{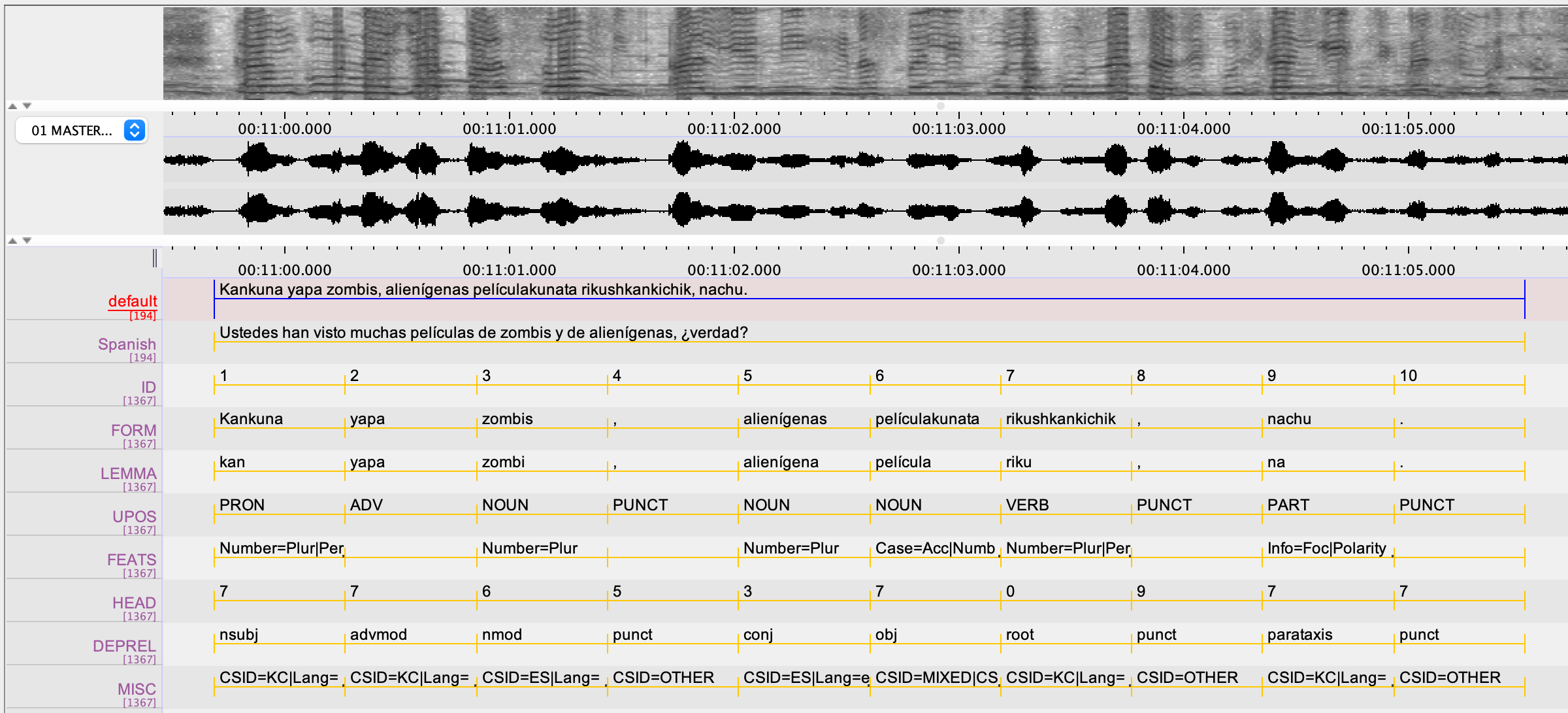}
    \caption{A screenshot of annotating a transcription, its Spanish translation, and UD-style morphosyntactic information in ELAN.}
    \label{fig:elan-screenshot}
\end{figure*}

The annotation of the dataset contains the following elements:
time-aligned sentence-level transcriptions, their translation in Spanish, and morphosyntactic annotation compatible with UD.
All of these annotations were done in ELAN \citep{elan}, and the annotated data are saved as XML-based EAF (ELAN Annotation Format) files.
ELAN is software commonly used to annotate spoken audio and video clips collected during linguistic fieldwork.
A screenshot of the annotation interface for the dataset building in this study is shown in Figure \ref{fig:elan-screenshot}.
To create an ASR dataset containing pairs of an audio sample and its transcription, the original audio files were segmented into sentence-level audio files based on the timestamps logged in the EAF files.
To process the annotation document with Python, the annotated EAF files were parsed into Python objects by the \texttt{pympi} library \citep{pympi-1.70}.
Though there are UD treebanks that were converted from ELAN-native annotation \citep{ostling-etal-2017-universal}, our dataset is the first attempt to directly incorporate UD annotation in the CoNLL-U format into ELAN to our knowledge.

\subsubsection{Transcription}
The source website provides transcriptions in Kichwa for each episode.
However, there were three problems in using the provided transcriptions.
First, words that are actually said by the actors often differ from the transcriptions (token-level inconsistencies).
Second, the actors often insert short sentences or interjections that do not appear in the transcriptions (utterance-level inconsistencies).
Third, the provided transcriptions have inconsistencies in the orthography (orthographic inconsistencies).
Table \ref{tab:tr_acc} summarizes the errors that the original transcription had compared to manually corrected transcriptions.
The metrics used in the Table, Character Error Rate (CER), Word Error Rate (WER), and Word Information Lost (WIL), are defined as follows:
$$
\text{CER, WER} = \dfrac{S + D + I}{N}
$$
$$
\text{WIL} = 1 - \dfrac{C}{N} + \dfrac{C}{P},
$$
where $S$, $D$, $I$ are the numbers of necessary substitutions, deletions, and insertions, respectively, to match the reference text, and $N$ is the number of characters (for CER) or words (for WER and WIL).
$P$ is the number of words in the prediction, and $C$ is the number of correctly predicted words.
As the Table shows, the original transcription had 22.7\%~CER compared to the corrected transcriptions, meaning that approximately one in five characters was either missing, wrong, or unnecessary, and 54.6\%~WER, meaning that more than half of the originally transcribed words required some correction.
Furthermore, 7.2\% of the actual utterances was missing from the original transcriptions.
Because these discrepancies make it difficult to automatically align the transcriptions with the audio segments, every sentence was manually checked and aligned.

\npdecimalsign{.}
\nprounddigits{2}
\begin{table}[t]
    \centering
    \begin{tabular}{lN{2}{2}N{2}{2}} \toprule
        Metrics & Raw~(\%) & {Normalized}~(\%) \\ \midrule
        CER & 22.703902781023605 & 20.761371928971054 \\
        WER & 54.61139896373057 & 43.626943005181346 \\
        WIL & 74.13255940585722 & 55.85987835097995 \\ \midrule
        Empty ratio & 7.216494845360824 &  \\
        \bottomrule
    \end{tabular}
    \caption{A summary on how correct the original transcriptions are with respect to what is actually said in the audio.
    The ``Raw'' column shows a comparison with no preprocessing to the transcriptions, while the ``Normalized'' column shows the results after applying lowercasing and removing punctuation.
    ``Empty ratio'' refers to the ratio of the number of uttered sentences that were not in the original transcriptions out of the total number of sentences.
    }
    \label{tab:tr_acc}
\end{table}
\npnoround

\subsubsection{Translation}
The Spanish translations of the transcriptions are also given by Radialistas.
However, they tend to be free translations that depend on the surrounding contexts and sometimes deviate from the information expressed in Kichwa.
For this reason, the translations were manually checked by a Kichwa--Spanish bilingual speaker and were corrected if necessary.

\subsubsection{Morphosyntactic annotation}
The morphosyntactic annotation of this dataset follows the CoNLL-U format of UD that annotates the lemma (\texttt{LEMMA}), part-of-speech (\texttt{UPOS}), morphological features (\texttt{FEATS}), syntactic head (\texttt{HEAD}), and dependency relation (\texttt{DEPREL}) for each token.

\begin{table*}[t]
    \centering
    \begin{tabular}{lll} \toprule
        Feature & Morpheme & Description \\ \midrule
        \texttt{Deixis=Ven} & \textit{mu} & Ventive (cislocative) \\
        \texttt{Focus=Addit} & \textit{pash}, \textit{pish} & Additive focus-sensitive marker \\
        \texttt{Focus=Restr} & \textit{lla} & Restrictive focus-sensitive marker \\
        \texttt{Info=Top} & \textit{ka} & Topic in the information structure \\
        \texttt{Info=Foc} & \textit{mi}, \textit{chu}, \textit{tak} & Focus in the information structure \\
        \texttt{State=Cont} & \textit{rak} & Continuative state \\
        \texttt{Switch=Same} & \textit{shpa}, \textit{nkapak} & Switch reference with the same subject \\
        \texttt{Switch=Diff} & \textit{kpi}, \textit{chun} & Switch reference with a different subject \\
        \bottomrule
    \end{tabular}
    \caption{A list of newly introduced morphological features.}
    \label{tab:new_feats}
\end{table*}

Since Kichwa is highly agglutinative and employs a number of suffixes to express functional meanings, there are several morphological features that are absent in the standard UD guidelines and are newly introduced in this dataset.
A list of newly introduced morphological features are summarized in Table \ref{tab:new_feats}.
The feature \texttt{Deixis=Ven} stands for the ventive (cislocative) morpheme that expresses the ``coming'' motion in the action expressed by the verb.
The feature key \texttt{Focus=} corresponds to focus-sensitive morphemes.
Kichwa has the additive focus marker \textit{=pash} ``also'' and the restrictive focus marker \textit{=lla} ``only''.
The feature key \texttt{Switch=} is used to mark the switch-reference features \citep{finer-1985} that co-occur with converbs\footnote{A converb is ``a nonfinite verb form whose main function is to mark
adverbial subordination'' \citep{haspelmath-1995}.}.
Switch-reference in Kichwa specifies whether the subject of the subordinate clause is the same as or different from that of the main clause.
For example, in (\ref{samesubject}), the subject is the same, while in (\ref{diffsubject}), the subject is different:

\begin{covsubexamples}
    \item\label{samesubject} \digloss[]
    {miku-nkapak muna-ni.}
    {eat-\textsc{cnv.prp.\textbf{ss}} want-\textsc{prs.1sg}}
    {I want to eat.}

    \item\label{diffsubject} \digloss[]
    {miku-chun muna-ni.}
    {eat-\textsc{cnv.prp.\textbf{ds}} want-\textsc{prs.1sg}}
    {I want (somebody else) to eat.\footnote{\textsc{cnv}: converb, \textsc{prp}: purposive mood, \textsc{ss}: same subject, \textsc{ds}: different subject}}
\end{covsubexamples}

Another significant modification from the standard UD guidelines is that this dataset annotates topic and focus in Kichwa as morphological features.
In current UD, morphological features cannot express grammatically marked topic and focus, because the guidelines do not have any features for them.
One reason for this treatment is that markers like topic and focus are syntactically less selective and can be attached to both nominal and verbal expressions.
Because UD's morphological features only allow for lexical, nominal (e.g., case), and verbal features (e.g., tense), features that have to do with the information structure cannot fit in the framework.
Indeed, unlike canonical affixes, the morphemes listed in Table \ref{tab:new_feats} have less syntactic restrictions as to which syntactic category they can be attached to; therefore, previous studies call them \textit{morfemas independientes} ``independent morphemes'' \citep{chasiquiza-2019} or \textit{enclíticos} ``enclitics'' \citep{catta-1994}.

In the standard UD guidelines, clitics are usually treated as independent tokens and are not represented as morphological features of the head token.
However, in this approach, it is impossible to annotate the topic and focus features as morphological features, and therefore information structure remains underrepresented in current UD for topic-prominent languages \citep{li-1976} that mark topic and focus morphologically like Kichwa.
For this reason, we tentatively added those information-structural features, which can be automatically converted to separate tokens if necessary.

Given the frequent code-mixing with Spanish in spoken Kichwa, the annotation in the dataset also includes the language code and the intra-word code-switching boundary for each token.
The code-switching annotation is listed in the \texttt{MISC} column, following the format in other code-switching UD treebanks \citep{cetinoglu-2022}.

\subsection{Analysis}
\begin{table}[t]
    \setlength{\tabcolsep}{2.5pt}
    \sisetup{round-mode=places, detect-all}
    \centering
    \begin{tabular}{lSll} \toprule
        Measure & {Kichwa} & min & max \\ \midrule
        TTR & 0.24 & 0.17 (vi) & 0.59 (kor.kai) \\
        MSP & \bfseries 3.73 & 0.99 (kor.kai) & 2.52 (chu) \\
        WS & \bfseries 1.05 & 0.16 (urd) & 0.62 (lat.itt) \\
        WH & 10.70 & 8.94 (afr) & 12.84 (kor.gsd) \\
        LH & 8.11 & 7.99 (chu) & 12.85 (kor.gsd) \\
        IS & \bfseries 34.31 & 0.00 (jpn) & 19.13 (eus) \\
        MFH & \bfseries 5.20 & 1.03 (kor.gsd) & 4.04 (ces.fic) \\
        \bottomrule
    \end{tabular}
    \caption{The morphological complexity scores of our Kichwa dataset and its comparison to the minimum and maximum scores reported in \citet{coltekin-2022}.
    The codes in parentheses refer to specific UD datasets, and the measures are type--token ratio (TTR), mean size of paradigm (MSP), information in word structure (WS), word entropy (WH), lemma entropy (LH), inflectional synthesis (IS), and morphological feature entropy (MFH); see \citet{coltekin-2022} for details.
    The boldfaced scores in Kichwa mean that they are higher than any other reported scores.}
    \label{tab:morph_complexity}
\end{table}

This subsection provides a brief analysis of our dataset with a focus on agglutinativity and code-switching of Kichwa.

\paragraph{Morphological complexity.}
Table~\ref{tab:morph_complexity} reports the morphological complexity scores of our Kichwa dataset based on the measures proposed in \citet{coltekin-2022}.
The table demonstrates that the morphological complexity of the Kichwa dataset is the highest for MSP (mean size of paradigm), WS (information in word structure), IS (inflectional synthesis), MFH (morphological feature entropy).
This shows the extremely high agglutinativity of Kichwa morphology, because MSP, IS, and MFH are calculated based on the diversity of inflected forms and morphological features.
On the other hand, our dataset did not show a high degree of complexity in terms of TTR (type--token ratio), WH (word entropy), and LH (lemma entropy).
This implies that there is not much diversity in the vocabulary of the dataset, since the dataset consists of a series of stories and has common topics and characters throughout the radio program.

\paragraph{Code-switching.}
Table \ref{tab:cs_stats} shows the distribution of languages in the dataset.
Code-switched tokens comprise $\sim$11.19\% of the entire dataset, and approximately half of them are word-internally code-switched tokens.
It is empirically known that agglutinative languages in language contact tend to derive morpheme-level code-switching such as in Turkish--German \citep{cetinoglu-2019}, and the code-switching distribution in Kichwa also follows this tendency.
As \citet{deibel-2019} pointed out, code-switched Spanish words appear either in an uninflected form (root) or in a fully inflected form, which can be followed by Kichwa morphemes, and, on the contrary, Kichwa stems are not followed by Spanish morphemes.
In terms of the selectivity of parts-of-speech, various syntactic categories can be code-switched.
Though open-class categories such as nouns and verbs commonly exhibit code-switching, closed-class categories like conjunctions also employ Spanish words, particularly in spoken varieties, as exemplified in the underlined word in (\ref{conj_cs}).
Other colored segments are open-class Spanish words.

\begin{covexample}\label{conj_cs}
    \digloss[]
    {kay=ka \textcolor{myGreen}{gasto}=chu \uline{\textcolor{myGreen}{o}} \textcolor{myGreen}{inversión}=chu ka-n.}
    {this=\textsc{top} expense=\textsc{foc.plq} or investment=\textsc{foc.plq} be-\textsc{prs.3}}
    {Are these expenses or investments?\footnote{\textsc{top}: topic, \textsc{plq}: polar question}}
\end{covexample}

\npdecimalsign{.}
\nprounddigits{2}
\begin{table}[t]
    \centering
    \begin{tabular}{llN{2}{2}} \toprule
        Language & & {Ratio~(\%)} \\ \midrule
        Kichwa-only & & 64.14622122786394 \\ \cmidrule{2-3}
        \multirow{2}{*}{Code-switched} & Spanish-only & 5.82837207538584 \\
        & Spanish--Kichwa & 5.5932653293390464 \\
        \bottomrule
    \end{tabular}
    \caption{A summary of the ratios of code-switched tokens.
    `Spanish--Kichwa` shows the ratio of tokens with intra-word code-switching.
    Other tokens are punctuation symbols.}
    \label{tab:cs_stats}
\end{table}
\npnoround

\section{Experiments: ASR}\label{sec:experiments}
This section reports the results of training the first ASR models for Kichwa based on our proposed dataset.

\subsection{Setup}
We developed a Kichwa ASR model by fine-tuning \texttt{wav2vec2-xlsr-53} with the Kichwa dataset.
Wav2Vec2 is a framework for pretraining a self-supervised ASR model that learns contextualized speech representations \citep{baevski2020wav2vec}.
Wav2Vec2 first segments the raw speech input into frames with 16kHz sampling rate and encodes into 512-dimensional features through 7 convolution blocks.
The feature vectors are then quantized into discrete values using Gumbel-softmax \citep{jang2017categorical}, and these quantized values are used as the labels later during the pretraining.
For the training step, some parts of the input values are masked, and the same feature vectors are fed into Transformer layers to predict the discrete labels of the masked frames, through which the model learns generalized speech representations.
In this way, Wav2Vec2 does not require manually labeled datasets for training and is able to be flexibly fine-tuned to a wide range of speech-related downstream tasks.
In particular, its offset pretrained model Wav2Vec2-XLSR-53 is trained on 53 languages, and it has been empirically shown that it has a strong adaptability to various languages by fine-tuning with small datasets \citep{conneau2020unsupervised}.

For our purpose, the training, validation, and test sets were generated by an 8:1:1 split, respectively.
During the preprocessing, we removed samples shorter than 1 second and longer than 15 seconds to ensure that frame masking is correctly done and to prevent the out-of-memory error, respectively.
The learning rate was set to $10^{-4}$.
We also trained models with smaller training sizes with 500, 1k, and 2k samples to imitate various degrees of low-resource settings.
The training was run for 30 epochs on 1 NVIDIA A10 GPU with a 24GB RAM.
The training took about 6 hours to complete, and the average power usage during the training was about 102W.
For the evaluation metrics, we used CER, WER, and WIL.

\subsection{Results}
\begin{table}[t]
    \centering
    \sisetup{round-mode=places, detect-all}
    \begin{tabular}{lS[round-precision=2]S[round-precision=2]S[round-precision=2]} \toprule
         & {CER} & {WER} & {WIL} \\ \midrule
         \textsc{Killkan}, all & \bfseries 2.9428604145196383 & \bfseries 19.938335046248715 & \bfseries 34.2922568321228 \\
         \textsc{Killkan}, 2k & 3.566930035497538 & 23.278520041109968 & 39.301037788391113 \\
         \textsc{Killkan}, 1k & 4.963929920989351 & 32.579650565262075 & 51.81505680084229 \\
         \textsc{Killkan}, 500 &  7.351425626932326 & 47.12230215827338 & 69.31318640708923 \\ \midrule
         Huqariq & 28.73 & {---} & {---} \\
         \bottomrule
    \end{tabular}
    \caption{The results of Kichwa ASR on the test set.
    Huqariq \citep{zevallos-etal-2022-huqariq} shows the result of Southern Quechua ASR fine-tuned on 144-hour data with pretrained Spanish Wav2Vec2.
    Note that their results are from their test set in Southern Quechua and not from our test set in Kichwa.
    }
    \label{tab:asr_results}
\end{table}

\begin{table*}
    \centering
    \begin{tabular}{lll} \toprule
        \multirow{2}{*}{Spanish} & Reference & Shuk \textbf{periodista ecuatoriano} rurashka. \\
        & Prediction & Shuk \textbf{periodiste cuatoriano} rurashka. \\ \midrule
        \multirow{2}{*}{Code-switching} & Reference & Ama kayapa alcaldíata \textbf{visita}nkapak sakishun. \\ 
        & Prediction & Ama kayapa alcaldíata \textbf{wisita}nkapak sakishun. \\ \midrule
        \multirow{2}{*}{Kichwa} & Reference & Ñukanchikpa tarpushkataka yalli \textbf{mishki} kan. \\
        & Prediction & Ñukanchikpa tarpushkataka yalli \textbf{nishki} kan. \\ \midrule
        \multirow{2}{*}{Spacing} & Reference & \textbf{Shuk kalluka} rurashallami ninkapak. \\
        & Prediction & \textbf{Shukkalluka} rurashallami ninkapak. \\ \midrule
        \multirow{2}{*}{Punctuation} & Reference & Mana pitapash llakichik\textbf{?} \\
        & Prediction & Mana pitapash llakichik\textbf{.} \\ \midrule
        \multirow{2}{*}{Alternative spelling} & Reference & \textbf{Kikinpak} warmi muspa ñawi mana pinkay niwarka.\\
        & Prediction & \textbf{Kikinpa} warmi muspa, ñawi mana pinkay niwarka. \\ \midrule
        \multirow{2}{*}{Interjection} & Reference & Paykunapa kawsaykunaka, \textbf{uff}, ninan llakipimi kan. \\
        & Prediction & Paykunapa kawsaykunaka, ninan llakipimi kan. \\
        \bottomrule
    \end{tabular}
    \label{tab:output_example}
    \caption{Examples of errors in the predicted transcriptions for the dev set.
    Errors are in bold-faced type.}
\end{table*}

The experimental results are shown in Table \ref{tab:asr_results}.
It compares four ASR models trained on different numbers of training samples: all samples~(\totaltrainsamples), 2k samples, 1k samples, and 500 samples, with the same hyperparameters.
The best model was the one trained with the most training data, which conforms with the general trend in machine learning.

For comparison, Table \ref{tab:asr_results} also lists the CER score of the Southern Quechua ASR model (Huqariq) reported in \citet{zevallos-etal-2022-huqariq}; Huqariq was fine-tuned on Spanish monolingual Wav2Vec2 with 144-hour Southern Kichwa training data.
Though the test datasets and the pretrained models are different between our studies and theirs, the clear contrast in CER (28.73\% and 2.94\%) shows the relatively successful performance of the Kichwa ASR model that was only trained on less than 3\%~of the Southern Quechua training data.
Importantly, even the extremely low-resource scenario with only 500 training samples achieved 7.35\%~CER.
Note that WER in Kichwa can be higher than WER in analytic languages like English, as tokens in Kichwa tend to consist of more characters with multiple agglutinated suffixes.
For example, the average length of English tokens in the GUM corpus \citep{Zeldes2017} is 4.08 while that of Kichwa tokens in our dataset is 6.04.

\subsection{Error analysis}
\begin{table}[t]
    \centering
    \sisetup{round-mode=places, detect-all}
    \begin{tabular}{lS[round-precision=2]} \toprule
        Error & {Ratio~(\%)} \\ \midrule
        \texttt{Punctuation} & 34.32343234 \\
        \texttt{Kichwa} & 27.39273927 \\
        \texttt{Alternative spelling} & 12.21122112 \\
        \texttt{Spanish} & 10.2310231 \\
        \texttt{Code-switching} & 10.2310231 \\
        \texttt{Spacing} & 4.95049505 \\
        \texttt{Interjection} & 0.6600660066 \\ \bottomrule
    \end{tabular}
    \caption{The distribution of each transcription error type in the dev set.}
    \label{tab:error_dist}
\end{table}

For an error analysis, we prepared seven error types (\texttt{Spanish}, \texttt{Code-switching}, \texttt{Kichwa}, \texttt{Spacing}, \texttt{Punctuation}, \texttt{Alternative spelling}, \texttt{Interjection}) and categorized the errors found in the dev set.
\texttt{Spanish}, \texttt{Code-switching}, and \texttt{Kichwa} are errors in transcribing tokens in those languages.
\texttt{Spacing} is an error where unnecessary spacing is inserted or a necessary spacing is omitted.
\texttt{Punctuation} is an error in choosing a punctuation symbol or capitalization.
\texttt{Alternative spelling} is an error where the spellings in both the reference and prediction texts are acceptable.
In other words, this type of error is not a wrong transcription in practice.
\texttt{Interjection} is an error in transcribing an interjection tokens.
Table \ref{tab:output_example} lists an actual prediction given by the model for each error type.

Table \ref{tab:error_dist} provides the distribution of the transcription error types found in the dev set.
The most common errors were punctuation errors, which took up more than one-third of the errors.
Given the fact that 67.81\% of the dataset is Kichwa tokens and 8.19\% either Spanish or code-switched as shown in Table \ref{tab:cs_stats}, it can be observed that Spanish and code-switching tokens tend to cause errors relatively more often than Kichwa tokens.
Because code-switched Spanish words tend to be either technical words, proper nouns, or other relatively uncommon words, it is difficult to train the model to be able to predict such corner cases correctly in this monolingual fine-tuning.
The prediction examples also exhibit the model's confusion in different Spanish spellings that share the same phoneme, such as <ho>/<o> and <v>/<b>.
Investigation of the methods to improve the transcription of low-frequency code-switched segments is beyond the scope of this study and is left for future work.

\begin{table}[t]
    \centering
    \sisetup{round-mode=places, detect-all}
    \begin{tabular}{lS[round-precision=2]S[round-precision=2]S[round-precision=2]} \toprule
         & {CER} & {WER} & {WIL} \\ \midrule
         \textsc{Killkan}, all & \bfseries 2.0425127657047855 & \bfseries 13.41212744090442 & \bfseries 23.26834797859192 \\
         \textsc{Killkan}, 2k & 2.6897043106519417 & 16.95786228160329 & 29.030781984329224 \\
         \textsc{Killkan}, 1k & 3.7881486759292246 & 24.20349434737924 & 39.44777846336365 \\
         \textsc{Killkan}, 500 & 6.127538297114357 & 39.26002055498458 & 59.98430252075195 \\
         \bottomrule
    \end{tabular}
    \caption{The results of Kichwa ASR on the test set after normalizing texts by lowercasing and removing punctuation.
    }
    \label{tab:asr_results_norm}
\end{table}

Considering the fact that the most common errors were mere punctuation errors and capitalization errors, we also measured the metrics after normalizing texts by lowercasing and removing punctuation.
As summarized in Table \ref{tab:asr_results_norm}, without casing and punctuation errors, CER was 2.04\%~and WER 13.41\%~for the best performing model.

\section{Conclusion}\label{sec:conclusion}
This study presented \textsc{Killkan}, the first linguistic dataset for Kichwa.
It contains speech and manually annotated transcription, Spanish translation, and morphosyntactic parsing information in UD's CoNLL-U format.
Our dataset also annotates morpheme-level code-switching with Spanish, which enabled us to conduct linguistic analyses related to code-switching such as measuring code-switching frequency.

Our study showcased the process of resource building and ASR model development for an extremely low-resource language.
The experimental results demonstrated 2.04\%~CER for the speech recognition task by the ASR model trained on less than 4 hours of audio data from our \textsc{Killkan} dataset.
Though this is a promising result for the extremely low-resource language, the analysis of the predicted output highlighted the difficulty for the model to correctly predict uncommon code-switched words.
Since code-switching is a common linguistic activity found across all over the world, especially among endangered languages in contact with other prestige languages, it is an important remaining task to improve prediction accuracy of code-switched words.
Also, the experimental results suggested that having more training samples is likely to contribute to improving the performance of Kichwa ASR, calling for more active resource building for low-resource languages.

\section{Ethical Considerations}
As our dataset has been developed only based on publicly available audio data, there is no direct concern of copyright infringement in this work.
However, there are several potential ethical concerns pertaining to technologies for low-resource languages in general.

\paragraph{Accessibility.}
Though our dataset and model are publicly available, the mode of the distribution is primarily in English, which might be an obstacle for the non-English-speaking users.
We will try to mitigate the disproportionate accessibility by adding descriptions in Kichwa and Spanish.

\paragraph{Demand by the community.}
Although our project was positively regarded by several native speakers during the first author's fieldwork in Quito, it does not mean that the technology should be embraced unconditionally by all speakers.

\paragraph{Language standardization.}
As described in Section~\ref{sec:background}, Ecuadorian Kichwa has a number of subdialects that have slightly different vocabulary, phonology, and morphology from each other.
Since our dataset and our ASR model are based on the standardized writing system, they might become an implicit force to use linguistic expressions of standardized Kichwa.
While this could be a positive effect on the literacy, it could also negatively affect the linguistic diversity of the Kichwa-speaking world.

\section{Acknowledgements}
This material is based upon work supported by the National Science Foundation under Grant No.~BCS-2109709 and by the University of Notre Dame under the Summer Language Abroad Grant (Quechua).
We are grateful to Luis Santill{\'a}n and Lourdes Perugachi for the feedback on the project and the support during the linguistic fieldwork.
We also thank the feedback given at the VI Seminario Internacional Revitalizando Ando and at the Tecnolog{\'i}as Digitales y Lenguas Ind{\'i}genas Workshop.

\section{Bibliographical References}\label{sec:reference}

\bibliographystyle{lrec-coling2024-natbib}
\bibliography{references}

\section{Language Resource References}
\label{lr:ref}
\bibliographystylelanguageresource{lrec-coling2024-natbib}
\bibliographylanguageresource{languageresource}

\section*{Appendix A. Glossing abbreviations}
\label{app:glossing}
\begin{table}[h]
    \centering
    \begin{tabular}{ll} \toprule
        Gloss & Function \\ \midrule
        \textsc{1, 2, 3} & 1st, 2nd, 3rd person \\
        \textsc{cnv} & converb \\
        \textsc{ds} & different subject \\
        \textsc{foc} & focus \\
        \textsc{plq}
        \textsc{prog} & progressive aspect \\
        \textsc{prosp} & prospective aspect \\
        \textsc{prp} & purposive mood \\
        \textsc{prs} & present tense \\
        \textsc{rcp} & reciprocal voice \\
        \textsc{refl} & reflexive voice \\
        \textsc{sg} & singular \\
        \textsc{ss} & same subject \\
        \textsc{ter} & terminative case \\
        \textsc{top} & topic \\
        \bottomrule
    \end{tabular}
    \caption{A list of glossing abbreviations used in the paper.}
    \label{tab:glossing}
\end{table}

\end{document}